# PRIORITY UNION AND GENERALIZATION IN DISCOURSE GRAMMARS

Claire Grover, Chris Brew, Suresh Manandhar, Marc Moens
HCRC Language Technology Group
*The University of Edinburgh*
2 Buccleuch Place
Edinburgh EH8 9LW, UK
Internet: C.Grover@ed.ac.uk

## Abstract

We describe an implementation in Carpenter's typed feature formalism, ALE, of a discourse grammar of the kind proposed by Scha, Polanyi, *et al*. We examine their method for resolving parallelism-dependent anaphora and show that there is a coherent feature-structural rendition of this type of grammar which uses the operations of *priority union* and *generalization*. We describe an augmentation of the ALE system to encompass these operations and we show that an appropriate choice of definition for priority union gives the desired multiple output for examples of VP-ellipsis which exhibit a strict/sloppy ambiguity.

## 1 Discourse Grammar

Working broadly within the sign-based paradigm exemplified by HPSG (Pollard and Sag in press) we have been exploring computational issues for a discourse level grammar by using the ALE system (Carpenter 1993) to implement a discourse grammar. Our central model of a discourse grammar is the Linguistic Discourse Model (LDM) most often associated with Scha, Polanyi, and their coworkers (Polanyi and Scha 1984, Scha and Polanyi 1988, Prüst 1992, and most recently in Prüst, Scha and van den Berg 1994). In LDM rules are defined which are, in a broad sense, unification grammar rules and which combine *discourse constituent units* (DCUs). These are simple clauses whose syntax and underresolved semantics have been determined by a sentence grammar but whose fully resolved final form can only be calculated by their integration into the current discourse and its context. The rules of the discourse grammar act to establish the rhetorical relations between constituents and to perform resolution of those anaphors whose interpretation can be seen as a function of discourse coherence (as opposed to those whose interpretation relies on general knowledge).

For illustrative purposes, we focus here on Prüst's rules for building one particular type of rhetorical relation, labelled "*list*" (Prüst 1992). His central thesis is that for DCUs to be combined into a *list* they must exhibit a degree of syntactic-semantic parallelism and that this parallelism will strongly determine the way in which some kinds of anaphor are resolved. The clearest example of this is VP-ellipsis as in (1a) but Prüst also claims that the subject and object pronouns in (1b) and (1c) are parallelism-dependent anaphors when they occur in *list* structures and must therefore be resolved to the corresponding fully referential subject/object in the first member of the *list*.

(1)     a. Hannah likes beetles. So does Thomas.
        b. Hannah likes beetles. She also likes caterpillars.
        c. Hannah likes beetles. Thomas hates them.

(2) is Prüst's *list* construction rule. It is intended to capture the idea that a *list* can be constructed out of two DCUs, combined by means of connectives such as *and* and *or*. The categories in Prüst's rules have features associated with them. In (2) these features are **sem** (the unresolved semantic interpretation of the category), **consem** (the contextually resolved semantic interpretation), and **schema** (the semantic information that is common between the daughter categories).

(2)   list [ **sem** : $C_1 \mathcal{R} ((C_1 \not\phi S_2) \sqcap S_2)$,
           **schema** : $C_1 \not\phi S_2$ ] $\longrightarrow$
    $DCU_1$ [ **sem** : $S_1$, **consem** : $C_1$ ] +
    $DCU_2$ [ **sem** : $\mathcal{R} S_2$, **consem** : $((C_1 \not\phi S_2) \sqcap S_2)$]
Conditions:
      $C_1 \not\phi S_2$ is a characteristic generalization of $C_1$
      and $S_2$; R $\in$ {and, or, ...}.

Prüst calls the operation used to calculate the value for **schema** the *most specific common denominator* (MSCD, indicated by the symbol $\not\phi$). The MSCD of $C_1$ and $S_2$ is defined as the most specific generalization of $C_1$ that can unify with $S_2$. It is essential that the result should be contentful to a degree that confirms that the *list* structure is an appropriate analysis, and to this end Prüst imposes the condition that the value of **schema** should be a *characteristic generalization* of the information contributed by the two daughters. There is no formal definition of this notion; it would require knowledge from many sources to determine whether sufficient informativeness had been achieved. However, assuming that this condition is met, Prüst uses the common information as a source for resolution of underspecified elements in the second daughter by encoding as the value of the second daughter's **consem** the unification of the result of MSCD with its pre-resolved semantics (the formula $((C_1 \not\sqsubseteq S_2) \sqcap S_2)$). So in Prüst's rule the MSCD operation plays two distinct roles, first as a test for parallelism (as the value of the mother's **schema**) and second as a basis for resolution (in the composite operation which is the value of the second daughter's **consem**). There are certain problems with MSCD which we claim stem from this attempt to use one operation for two purposes, and our primary concern is to find alternative means of achieving Prüst's intended analysis.

## 2 An ALE Discourse Grammar

For our initial exploration into using ALE for discourse grammars we have developed a small discourse grammar whose lexical items are complete sentences (to circumvent the need for a sentence grammar) and which represents the semantic content of sentences using feature structures of type *event* whose sub-types are indicated in the following part of the type hierarchy:

(3) 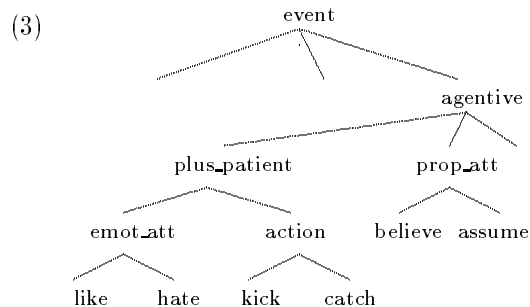

In addition we have a very simplified semantics of noun phrases where we encode them as of type *entity* with the subtypes indicated below:

(4) 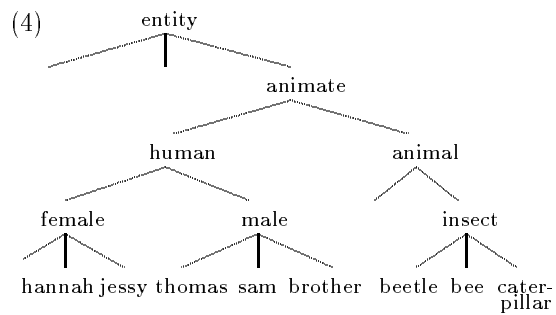

Specifications of which features are appropriate for which type give us the following representations of the semantic content of the discourse units in (1):

(5)  a.  *Hannah likes beetles*
$$\begin{bmatrix} \text{AGENT} & hannah \\ \text{PATIENT} & beetle \end{bmatrix}$$
$like$

b.  *So does Thomas*
$$\begin{bmatrix} \text{AGENT} & thomas \end{bmatrix}$$
$agentive$

c.  *She also likes caterpillars*
$$\begin{bmatrix} \text{AGENT} & female \\ \text{PATIENT} & caterpillar \end{bmatrix}$$
$like$

d.  *Thomas hates them*
$$\begin{bmatrix} \text{AGENT} & thomas \\ \text{PATIENT} & entity \end{bmatrix}$$
$hate$

### 2.1 Calculating Common Ground

The SCHEMA feature encodes the information that is common between daughter DCUs and Prüst uses MSCD to calculate this information. A feature-structural definition of MSCD would return as a result the most specific feature structure which is at least as general as its first argument but which is also unifiable with its second argument. For the example in (1c), the MSCD operation would be given the two arguments in (5a) and (5d), and (6) would be the result.

(6)
$$\begin{bmatrix} \text{AGENT} & human \\ \text{PATIENT} & beetle \end{bmatrix}$$
$emot\_att$

We can contrast the MSCD operation with an operation which is more commonly discussed in the context of feature-based unification systems, namely *generalization*. This takes two feature-structures as input and returns a feature structure which represents the common information in them. Unlike MSCD, generalization is not asymmetric, i.e. the order in which the arguments are presented does not affect the result. The generalization of (5a) and (5d) is shown in (7).

(7)
$$\begin{bmatrix} \text{AGENT} & human \\ \text{PATIENT} & entity \end{bmatrix}$$
$emot\_att$

It can be seen from this example that the MSCD result contains more information than the generalization result. Informally we can say that it seems to reflect the common information between the two inputs *after the parallelism-dependent anaphor in the second sentence has been resolved*. The reason it is safe to use MSCD in this context is precisely because its use in a *list* structure guarantees that the pronoun in the second sentence will be resolved to *beetle*. In fact the result of MSCD in this case is exactly the result we would get if we were to perform the generalization of the resolved sentences and, as a representation of what the two have in common, it does seem that this is more desirable than the generalization of the pre-resolved forms.

If we turn to other examples, however, we discover that MSCD does not always give the best results. The discourse in (8) must receive a constituent structure where the second and third clauses are combined to form a *contrast* pair and then this *contrast* pair combines with the first sentence to form a *list*. (Prüst has a separate rule to build *contrast* pairs but the use of MSCD is the same as in the *list* rule.)

(8)  Hannah likes ants. Thomas likes bees but Jessy hates them.

(9) 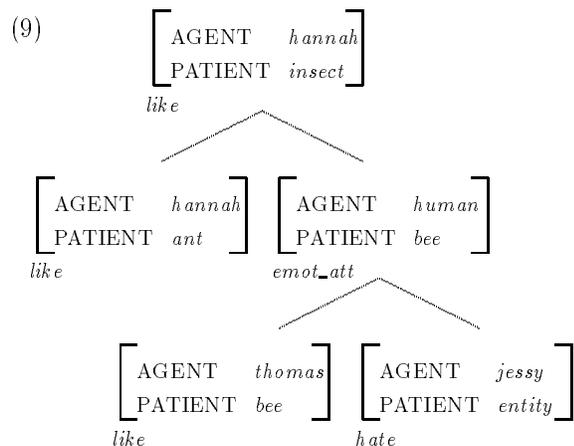

The tree in (9) demonstrates the required structure and also shows on the mother and intermediate nodes what the results of MSCD would be. As we can see, where elements of the first argument of MSCD are more specific than the corresponding elements in the second, then the more specific one occurs in the result. Here, this has the effect that the structure [*like*, AGENT *hannah*, PATIENT *insect* ] is somehow claimed to be common ground between all three constituents even though this is clearly not the case.

Our solution to this problem is to dispense with the MSCD operation and to use generalization instead. However, we do propose that generalization should take inputs whose parallelism dependent anaphors have already been resolved.[1] In the case of the combination of (5a) and (5d), this will give exactly the same result as MSCD gave (i.e. (6)), but for the example in (8) we will get different results, as the tree in (10) shows. (Notice that the representation of the third sentence is one where the anaphor is resolved.) The resulting generalization, [*emot_att*, AGENT *human*, PATIENT *insect*], is a much more plausible representation of the common information between the three DCUs than the results of MSCD.

(10) 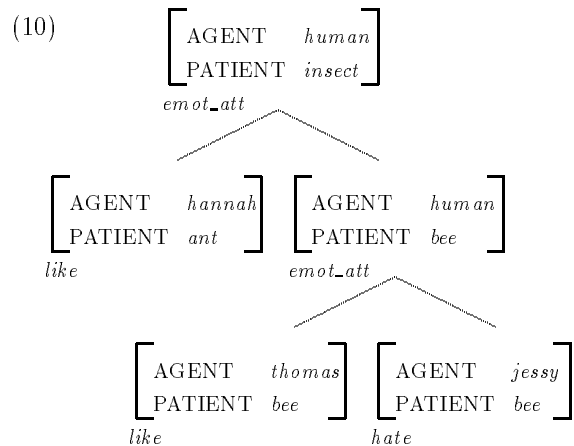

## 2.2 Resolution of Parallel Anaphors

We have said that MSCD plays two roles in Prüst's rules and we have shown how its function in calculating the value of SCHEMA can be better served by using the generalization operation instead. We turn now to the composite operation indicated in (2) by the formula $((C_1 \not\in S_2) \sqcap S_2)$. This composite operation calculates MSCD and then unifies it back in with the second of its arguments in order to resolve any parallelism-dependent anaphors that might occur in the second DCU. In the discussion that follows, we will refer to the first DCU in the *list* rule as the *source* and to the second DCU as the *target* (because it contains a parallelism-dependent anaphor which is the target of our attempt to resolve that anaphor).

In our ALE implementation we replace Prüst's composite operation by an operation which has occasionally been proposed as an addition to feature-based unification systems and which is usually referred to either as *default unification* or as *priority union*.[2] Assumptions about the exact definition of this operation vary but an intuitive description of it is that it is an operation which takes two feature structures and produces a result which is a merge of the information in the two inputs. However, the information in one of the feature structures is "strict" and cannot be lost or overridden while the information in the other is defeasible. The operation is a kind of union where the information in the strict structure takes priority over that in the default structure, hence our preference to refer to it by the name priority union. Below we demonstrate the results of priority union for the examples in (1a)–(1c). Note that the target is the strict structure and the source is the defeasible one.

---

[1] As described in the next section, we use priority union to resolve these anaphors in both *lists* and *contrasts*. The use of generalization as a step towards checking that there is sufficient common ground is subsequent to the use of priority union as the resolution mechanism.

[2] See, for example, Bouma (1990), Calder (1990), Carpenter (1994), Kaplan (1987).

(11) *Hannah likes beetles. So does Thomas.*

Source: 5a
Target: 5b
Priority Union:
$$\begin{bmatrix} \text{AGENT} & thomas \\ \text{PATIENT} & beetle \end{bmatrix}$$
$like$

(12) *Hannah likes beetles. She also likes caterpillars.*

Source: 5a
Target: 5c
Priority Union:
$$\begin{bmatrix} \text{AGENT} & hannah \\ \text{PATIENT} & caterpillar \end{bmatrix}$$
$like$

(13) *Hannah likes beetles. Thomas hates them.*

Source: 5a
Target: 5d
Priority Union:
$$\begin{bmatrix} \text{AGENT} & thomas \\ \text{PATIENT} & beetle \end{bmatrix}$$
$hate$

For these examples priority union gives us exactly the same results as Prüst's composite operation. We use a definition of priority union provided by Carpenter (1994) (although note that his name for the operation is "credulous default unification"). It is discussed in more detail in Section 3. The priority union of a target $T$ and a source $S$ is defined as a two step process: first calculate a maximal feature structure $S'$ such that $S' \sqsubseteq S$, and then unify the new feature structure with $T$.

This is very similar to Prüst's composite operation but there is a significant difference, however. For Prüst there is a requirement that there should always be a unique MSCD since he also uses MSCD to calculate the common ground as a test for parallelism and there must only be one result for that purpose. By contrast, we have taken Carpenter's definition of credulous default unification and this can return more than one result. We have strong reasons for choosing this definition even though Carpenter does define a "skeptical default unification" operation which returns only one result. Our reasons for preferring the credulous version arise from examples of VP-ellipsis which exhibit an ambiguity whereby both a "strict" and a "sloppy" reading are possible. For example, the second sentence in (14) has two possible readings which can be glossed as "Hannah likes Jessy's brother" (the strict reading) and "Hannah likes her own brother" (the sloppy reading).

(14) Jessy likes her brother. So does Hannah.

The situations where the credulous version of the operation will return more than one result arise from structure sharing in the defeasible feature structure and it turns out that these are exactly the places where we would need to get more than one result in order to get the strict/sloppy ambiguities. We illustrate below:

(15) *Jessy likes her brother. So does Hannah.*

Source:
$$\begin{bmatrix} \text{AGENT} & \boxed{1}\,jessy \\ \text{PATIENT} & \begin{bmatrix} \text{BROTHER-OF} & \boxed{1} \end{bmatrix} \\ & brother \end{bmatrix}$$
$like$

Target:
$$\begin{bmatrix} \text{AGENT} & hannah \end{bmatrix}$$
$agentive$

Priority Union:
$$\begin{bmatrix} \text{AGENT} & \boxed{1}\,hannah \\ \text{PATIENT} & \begin{bmatrix} \text{BROTHER-OF} & \boxed{1} \end{bmatrix} \\ & brother \end{bmatrix}$$
$like$

$$\begin{bmatrix} \text{AGENT} & hannah \\ \text{PATIENT} & \begin{bmatrix} \text{BROTHER-OF} & jessy \end{bmatrix} \\ & brother \end{bmatrix}$$
$like$

Here priority union returns two results, one where the structure-sharing information in the source has been preserved and one where it has not. As the example demonstrates, this gives the two readings required. By contrast, Carpenter's skeptical default unification operation and Prüst's composite operation return only one result.

### 2.3 Higher Order Unification

There are similarities between our implementation of Prüst's grammar and the account of VP-ellipsis described by Dalrymple, Shieber and Pereira (1991) (henceforth DSP). DSP gives an equational characterization of the problem of VP-ellipsis where the interpretation of the target phrase follows from an initial step of solving an equation with respect to the source phrase. If a function can be found such that applying that function to the source subject results in the source interpretation, then an application of that function to the target subject will yield the resolved interpretation for the target. The method for solving such equations is "higher order unification". (16) shows all the components of the interpretation of the example in (11).

(16) *Hannah likes beetles. So does Thomas.*

Source: $like(\underline{hannah}, beetle)$
Target (T): $P(thomas)$
Equation: $P(hannah) = like(\underline{hannah}, beetle)$
Solution: $P = \lambda x.like(x, beetle)$
Apply to T: $like(thomas, beetle)$

A prerequisite to the DSP procedure is the establishment of parallelism between source and target and the identification of parallel subparts. For example, for (16) it is necessary both that the two clauses *Hannah likes beetles* and *So does Thomas* should be parallel and that the element *hannah* should be identified as a parallel element. DSP indicate parallel elements in the source by means of underlines as shown in (16). An underlined element in the source is termed a 'primary occurrence' and DSP place a constraint on solutions to equations requiring that primary occurrences be abstracted. Without the identification of *hannah* as a primary occurrence in (16), other equations deriving from the source might be possible, for example (17):

(17) a. $P(beetle) = like(hannah, beetle)$
     b. $P(like) = like(hannah, beetle)$

The DSP analysis of our strict/sloppy example in (14) is shown in (18). The ambiguity follows from the fact that there are two possible solutions to the equation on the source: the first solution involves abstraction of just the primary occurrence of *jessy*, while the second solution involves abstraction of both the primary and the secondary occurrences. When applied to the target these solutions yield the two different interpretations:

(18) *Jessy likes her brother. So does Hannah.*

Source: $like(\underline{jessy}, brother\text{-}of(jessy))$
Target: $P(hannah)$
Equation: $P(jessy) = like(\underline{jessy}, brother\text{-}of(jessy))$
Sol.1 (S1): $P = \lambda x.like(x, brother\text{-}of(jessy))$
Sol.2 (S2): $P = \lambda x.like(x, brother\text{-}of(x))$
Apply S1: $like(hannah, brother\text{-}of(jessy))$
Apply S2: $like(hannah, brother\text{-}of(hannah))$

DSP claim that a significant attribute of their account is that they can provide the two readings in strict/sloppy ambiguities without having to postulate ambiguity in the source. They claim this as a virtue which is matched by few other accounts of VP-ellipsis. We have shown here, however, that an account which uses priority union also has no need to treat the source as ambiguous.

Our results and DSP's also converge where the treatment of cascaded ellipsis is concerned. For the example in (19) both accounts find six readings although two of these are either extremely implausible or even impossible.

(19) John revised his paper before the teacher did, and Bill did too.

DSP consider ways of reducing the number of readings and, similarly, we are currently exploring a potential solution whereby some of the reentrancies in the source are required to be transmitted to the result of priority union.

There are also similarities between our account and the DSP account with respect to the establishment of parallelism. In the DSP analysis the determination of parallelism is separate from and a prerequisite to the resolution of ellipsis. However, they do not actually formulate how parallelism is to be determined. In our modification of Prüst's account we have taken the same step as DSP in that we separate out the part of the feature structure used to determine parallelism from the part used to resolve ellipsis. In the general spirit of Prüst's analysis, however, we have taken one step further down the line towards determining parallelism by postulating that calculating the generalization of the source and target is a first step towards showing that parallelism exists. The further condition that Prüst imposes, that the common ground should be a characteristic generalization, would conclude the establishment of parallelism. We are currently not able to define the notion of characteristic generalization, so like DSP we do not have enough in our theory to fully implement the parallelism requirement. In contrast to the DSP account, however, our feature structural approach does not involve us having to explicitly pair up the component parts of source and target, nor does it require us to distinguish primary from secondary occurrences.

## 2.4 Parallelism

In the DSP approach to VP-ellipsis and in our approach too, the emphasis has been on semantic parallelism. It has often been pointed out, however, that there can be an additional requirement of syntactic parallelism (see for example, Kehler 1993 and Asher 1993). Kehler (1993) provides a useful discussion of the issue and argues convincingly that whether syntactic parallelism is required depends on the coherence relation involved. As the examples in (20) and (21) demonstrate, semantic parallelism is sufficient to establish a relation like *contrast* but it is not sufficient for building a coherent *list*.

(20) The problem was looked into by John, but no-one else did.

(21) *This problem was looked into by John, and Bill did too.

For a *list* to be well-formed both syntactic and semantic parallelism are required:

(22) John looked into this problem, and Bill did too.

In the light of Kehler's claims, it would seem that a more far-reaching implementation of our priority union account would need to specify how the constraint of syntactic parallelism might be implemented for those constructions which require it. An HPSG-style sign, containing as it does all types of linguistic information within the same feature structure, would lend itself well to an account of syntactic parallelism. If we consider that the DTRS feature in the sign for the source clause contains the entire parse tree including the node for the VP which is the syntactic antecedent, then ways to bring together the source VP and the target begin to suggest themselves. We have at our disposal both unification to achieve re-entrancy and the option to use priority union over syntactic subparts of the sign. In the light of this, we are confident that it would be possible to articulate a more elaborate account of VP-ellipsis within our framework and that priority union would remain the operation of choice to achieve the resolution.

## 3 Extensions to ALE

In the previous sections we showed that Prüst's MSCD operation would more appropriately be replaced by the related operations of generalization and priority union. We have added generalization and priority union to the ALE system and in this section we discuss our implementation. We have provided the new operations as a complement to the definite clause component of ALE. We chose this route because we wanted to give the grammar writer explicit control of the point at which the operations were invoked. ALE adopts a simple PROLOG-like execution strategy rather than the more sophisticated control schemes of systems like CUF and TFS (Manandhar 1993). In principle it might be preferable to allow the very general deduction strategies which these other systems support, since they have the potential to support a more declarative style of grammar-writing. Unfortunately, priority union is a non-monotonic operation, and the consequences of embedding such operations in a system providing for flexible execution strategies are largely unexplored. At least at the outset it seems preferable to work within a framework in which the grammar writer is required to take some of the responsibility for the order in which operations are carried out. Ultimately we would hope that much of this load could be taken by the system, but as a tool for exploration ALE certainly suffices.

### 3.1 Priority Union in ALE

We use the following definition of priority union, based on Carpenter's definition of credulous default unification:

(23) $\text{punion}(T,S) = \{\text{unify}(T,S') \mid S' \sqsubseteq S$ is maximal such that $\text{unify}(T,S')$ is defined$\}$

punion(T,S) computes the priority union of T (target; the strict feature structure) with S (source; the defeasible feature structure). This definition relies on Moshier's (1988) definition of *atomic feature structures*, and on the technical result that any feature structure can be decomposed into a unification of a unique set of atomic feature structures. Our implementation is a simple proceduralization of Carpenter's declarative definition. First we decompose the default feature structure into a set of atomic feature structures, then we search for the maximal subsets required by the definition.

We illustrate our implementation of priority union in ALE with the example in (15): *Source* is the default input, and *Target* is the strict input. The hierarchy we assume is the same as shown in (3) and (4). Information about how features are associated with types is as follows:

- The type *agentive* introduces the feature AGENT with range type *human*.
- The type *plus-patient* introduces the feature PATIENT with range type *human*.
- The type *brother* introduces the feature BROTHER-OF with range type *human*.
- The types *jessy* and *hannah* introduce no features.

In order to show the decomposition into atomic feature structures we need a notation to represent paths and types. We show paths like this: PATIENT|BROTHER-OF and in order to stipulate that the PATIENT feature leads to a structure of type *brother*, we include type information in this way: (PATIENT/*brother*)|(BROTHER-OF/*human*). We introduce a special feature (*) to allow specification of the top level type of the structure. The structures in (15) decompose into the following atomic components.

(24) Default input:

| | |
|---|---|
| (AGENT/*jessy*) | (D1) |
| (PATIENT/*brother*)\|(BROTHER-OF/*jessy*) | (D2) |
| AGENT = PATIENT\|BROTHER-OF | (D3) |
| (*/*like*) | (D4) |

Strict input:

| | |
|---|---|
| (AGENT/*hannah*) | (S1) |
| (*/*agentive*) | (S2) |

Given the type hierarchy the expressions above expand to the following typed feature structures:

(25)

Default input:

$$\begin{bmatrix} \text{AGENT} & jessy \end{bmatrix}_{agentive} \quad (D1)$$

$$\begin{bmatrix} \text{AGENT} & human \\ \text{PATIENT} & \begin{bmatrix} \text{BROTHER-OF} & jessy \end{bmatrix}_{brother} \end{bmatrix}_{plus\text{-}patient} \quad (D2)$$

$$\begin{bmatrix} \text{AGENT} & \boxed{1}\,human \\ \text{PATIENT} & \begin{bmatrix} \text{BROTHER-OF} & \boxed{1} \end{bmatrix}_{brother} \end{bmatrix}_{plus\text{-}patient} \quad (D3)$$

$$\begin{bmatrix} \text{AGENT} & human \\ \text{PATIENT} & entity \end{bmatrix}_{like} \quad (D4)$$

Strict input:

$$\begin{bmatrix} \text{AGENT} & hannah \end{bmatrix}_{agentive} \quad (S1,S2)$$

We can now carry out the following steps in order to generate the priority union.

1. Add (D4) to the strict input. It cannot conflict.
2. Note that it is impossible to add (D1) to the strict input.
3. Non-deterministically add either (D2) or (D3) to the strict input.
4. Note that the results are maximal in each case because it is impossible to add both (D2) and (D3) without causing a clash between the disjoint atomic types *hannah* and *jessy*.
5. Assemble the results into feature structures. If we have added (D3) the result will be (26) and if we have added (D2) the result will be (27).

(26) Result 1:

$$\begin{bmatrix} \text{AGENT} & \boxed{1}\,hannah \\ \text{PATIENT} & \begin{bmatrix} \text{BROTHER-OF} & \boxed{1} \end{bmatrix}_{brother} \end{bmatrix}_{like}$$

(27) Result 2:

$$\begin{bmatrix} \text{AGENT} & hannah \\ \text{PATIENT} & \begin{bmatrix} \text{BROTHER-OF} & jessy \end{bmatrix}_{brother} \end{bmatrix}_{like}$$

In order to make this step-by-step description into an algorithm we have used a breadth-first search routine with the property that the largest sets are generated first. We collect answers in the order in which the search comes upon them and carry out subsumption checks to ensure that all the answers which will be returned are maximal. These checks reduce to checks on subset inclusion, which can be reasonably efficient with suitable set representations. Consistency checking is straightforward because the ALE system manages type information in a manner which is largely transparent to the user. Unification of ALE terms is defined in such a way that if adding a feature to a term results in a term of a new type, then the representation of the structure is specialized to reflect this. Since priority union is non-deterministic we will finish with a set of maximal consistent subsets. Each of these subsets can be converted directly into ALE terms using ALE's built-in predicate `add_to/5`. The resulting set of ALE terms is the (disjunctive) result of priority union.

In general we expect priority union to be a computationally expensive operation, since we cannot exclude pathological cases in which the system has to search an exponential number of subsets in the search for the maximal consistent elements which are required. In the light of this it is fortunate that our current discourse grammars do not require frequent use of priority union. Because of the inherent complexity of the task we have favoured correctness and clarity at the possible expense of efficiency. Once it becomes established that priority union is a useful operation we can begin to explore the possibilities for faster implementations.

## 3.2 Generalization in ALE

The abstract definition of *generalization* stipulates that the generalization of two categories is the largest category which subsumes both of them. Moshier (1988) has shown that generalization can be defined as the intersection of sets of atomic feature structures. In the previous section we outlined how an ALE term can be broken up into atomic feature structures. All that is now required is the set intersection operation with the addition that we also need to cater for the possibility that atomic types may have a consistent generalization.

1. For $P$ and $Q$ complex feature structures
   $Gen(P,Q) =_{df} \{Path : C \mid Path : A \in P$
   and $Path : B \in Q \}$ where $C$ is the most specific type which subsumes both $A$ and $B$.
2. For $A$ and $B$ atomic types $Gen(A, B) =_{df} C$ where $C$ is the most specific type which subsumes both $A$ and $B$.

In ALE there is always a unique type for the generalization. We have made a small extension to the ALE compiler to generate a table of type generalizations to assist in the (relatively) efficient computation of generalization. To illustrate, we show how the generalization of the two feature structures in (28) and (29) is calculated.

(28) *Hannah likes ants.*

$$\begin{bmatrix} \text{AGENT} & hannah \\ \text{PATIENT} & ant \end{bmatrix}_{like}$$

(29) *Jessy laughs.*

$$\begin{bmatrix} \text{AGENT} & jessy \end{bmatrix}_{laugh}$$

These decompose into the atomic components shown in (30) and (31) respectively.

(30)      (*/*like*)
         (AGENT/*hannah*)
         (PATIENT/*ant*)

(31)      (*/*laugh*)
         (AGENT/*jessy*)

These have only the AGENT path in common although with different values and therefore the generalization is the feature structure corresponding to this path but with the generalization of the atomic types *hannah* and *jessy* as value:

(32) $\begin{bmatrix} \text{AGENT} & female \end{bmatrix}_{agentive}$

## 4 Conclusion

In this paper we have reported on an implementation of a discourse grammar in a sign-based formalism, using Carpenter's Attribute Logic Engine (ALE). We extended the discourse grammar and ALE to incorporate the operations of priority union and generalization, operations which we use for resolving parallelism dependent anaphoric expressions. We also reported on a resolution mechanism for verb phrase ellipsis which yields sloppy and strict readings through priority union, and we claimed some advantages of this approach over the use of higher-order unification.

The outstanding unsolved problem is that of establishing parallelism. While we believe that generalization is an appropriate formal operation to assist in this, we still stand in dire need of a convincing criterion for judging whether the generalization of two categories is sufficiently informative to successfully establish parallelism.

## Acknowledgements

This work was supported by the EC-funded project LRE-61-062 "Towards a Declarative Theory of Discourse" and a longer version of the paper is available in Brew *et al* (1994). We have profited from discussions with Jo Calder, Dick Crouch, Joke Dorrepaal, Claire Gardent, Janet Hitzeman, David Millward and Hub Prüst. Andreas Schöter helped with the implementation work. The Human Communication Research Centre (HCRC) is supported by the Economic and Social Research Council (UK).